# Severity Classification of Chronic Obstructive Pulmonary Disease in Intensive Care Units: A Semi-Supervised Approach Using MIMIC-III Dataset

Akram Shojaei[1] and Mehdi Delrobaei[1,2*]

[1*]Faculty of Electrical Engineering, K. N. Toosi University of Technology, Tehran, Tehran, Iran.
[2]Department of Electrical and Computer Engineering, Western University, London, ON, Canada.

*Corresponding author(s). E-mail(s): delrobaei@kntu.ac.ir;
Contributing authors: akram.shojaeibagheini@email.kntu.ac.ir;

## Abstract

Chronic obstructive pulmonary disease (COPD) is a major global health concern, with accurate severity assessment crucial for effective management, especially in intensive care units (ICUs). This study presents a novel approach to COPD severity classification using machine learning algorithms applied to the MIMIC-III dataset. Our work presents a new application of the MIMIC-III dataset and contributes to the growing field of artificial intelligence in critical care medicine. We developed a model to classify COPD severity based on available ICU parameters, including blood gas measurements and vital signs. Our methodology incorporated semi-supervised learning techniques to leverage unlabeled data, enhancing model robustness. A random forest classifier demonstrated superior performance, achieving 92.51% accuracy and 0.98 ROC AUC distinguishing between mild-to-moderate and severe COPD cases. This approach offers a practical, accurate, and accessible tool for rapid COPD severity assessment in ICU settings, potentially improving clinical decision-making and patient outcomes. Future research should focus on external validation and integration into clinical decision support systems to enhance COPD management in the ICUs.

# 1 Introduction

CHRONIC obstructive pulmonary disease (COPD) is a common respiratory condition characterized by compromised air flow and persistent symptoms, including emphysema and chronic bronchitis [1]. Exposure to air pollution and smoking can lead to COPD, causing respiratory symptoms like lung damage, congestion, coughing, breathing difficulties, and fatigue. Although there is currently no known cure for this condition, these symptoms can be managed through lifestyle changes, pharmacological interventions, oxygen therapy, and rehabilitation programs [2]. The economic impact of COPD is substantial due to its high rates of hospitalization and mortality. By 2030, COPD is expected to be the third most common cause of death globally [3]. It is crucial to take immediate measures such as prevention, early identification, and collaborative healthcare strategies to address the growing impact of COPD on global healthcare systems and mortality rates [4], [5].

Integrating machine learning methods in COPD assessment shows promise in accurately evaluating disease severity, potentially revolutionizing personalized treatment strategies based on predictive analytics and patient-specific data. Exacerbations in COPD are defined as acute deteriorations that surpass normal variations and require medication adjustments [6]. The severity assessment is pivotal in guiding therapy, facilitating the prognosis, and predicting the likelihood of exacerbations.

Spirometry measures lung function by air expulsion and is essential in diagnosing and assessing the severity of COPD [7]-[9]. However, limitations in forced maneuvers may be problematic for patients with cardiovascular conditions. The inability to cooperate in critical cases may hinder accurate spirometry [10]. Alternatives such as blood gas analysis become vital in intensive care unit (ICU) settings, where critically ill patients cannot perform spirometry, ensuring comprehensive respiratory assessment despite spirometry's constraints. The arterial blood gas (ABG) test is essential for quickly assessing patients' respiratory status in the ICU by measuring their blood oxygen levels, carbon dioxide levels, and *pH* balance [11]. Utilizing automated analyzers facilitates quick results while avoiding patient exertion via arterial catheters. Unlike spirometry, ABG testing allows for swift assessment of critical COPD parameters without requiring physical examination [12]. ABG parameters are crucial for classifying COPD severity, predicting exacerbations, guiding therapy, and assessing outcomes [13]. Respiratory assessment is critical for COPD management in the ICU without causing strain on patients. Vital signs, such as heart rate, respiratory rate, and blood oxygen saturation, are important indicators when assessing chronic respiratory diseases, especially during exacerbations [6]. Recent studies have found accessible markers for exacerbation severity increase before, during, and after episodes [6], [14]. Resting heart rate is directly linked with the severity of COPD, regardless of treatment, and increases the risk of mortality [14]. Therefore, continuous monitoring of vital signs in the ICU can provide real-time insights into the status of chronic respiratory diseases, aiding in severity assessment and intervention [9]. Healthcare professionals can use these noninvasive indicators to develop timely and personalized management strategies for patients with chronic respiratory diseases, essential for addressing their evolving conditions.

The MIMIC-III (Medical Information Mart for Intensive Care) dataset [15], [16], a comprehensive clinical database, is predominantly utilized in research for predicting mortality rates and diagnosing various diseases. By performing a series of pre-processing steps, we employed this dataset to evaluate the severity of COPD. The pre-processing procedure involved cleaning, transforming, and preparing the data to ensure its suitability for our analysis. We then applied machine learning algorithms to the pre-processed data to classify the severity of COPD. This research aims to assess COPD severity in ICU patients using machine learning algorithms, incorporating blood gas parameters and vital signs. Our approach is expected to significantly improve the accuracy of disease severity assessments, enabling more personalized and effective treatment strategies for COPD patients.

## 2 Related Work

Numerous studies have utilized the MIMIC-III dataset, focusing on predicting mortality rates or disease outcomes. M. Asad Bin Hameed *et al.* developed a machine learning model to predict acute pancreatitis (AP) mortality by augmenting datasets from MIMIC-III, MIMIC-IV, and eICU [17]. They addressed data quality issues using iterative imputation and up-sampling methods. The random forest classifier, trained on augmented data, achieved the highest recall, demonstrating the model's potential for early AP detection and improved patient outcomes. Mucan Liu *et al.* introduced an explainable knowledge distillation method using XGBoost (XGB-KD) to enhance ICU mortality prediction [18]. By distilling knowledge from deep learning models into XGBoost, their approach improved predictive performance and provided clear explanations. Evaluated on the MIMIC-III dataset, XGB-KD outperformed traditional and state-of-the-art methods, ensuring accuracy and interpretability.

Yutao Dou *et al.*. proposed the ITFG (interpretable tree-based feature generation) model for early sepsis detection, addressing the sequential organ failure assessment (SOFA) limitations. Their model, enhanced by the semi-supervised attention-based conditional transfer learning (SAC-TL) framework, achieved AUC scores of 97.98% on MIMIC and 86.21% on PhysioNet, demonstrating high effectiveness and adaptability in different data environments [19]. Sarika R. Khope *et al.* developed a novel predictive scheme using the MIMIC-III dataset to enhance prognosis in critical care [20]. Their study reviews various predictive models and clinical diagnostics, highlighting the effectiveness and limitations of existing methods. By leveraging advanced analytics and machine learning, their work comprehensively evaluates predictive schemes for improved patient outcomes and treatment planning. Numerous studies have been conducted to evaluate the severity of COPD.

Some studies have focused on categorizing computed tomography (CT) images [21]. Peng *et al.* used a multi-scale residual network to analyze raw CT images and their differential excitation components [22]. Their method accurately measured emphysema with 93.74% accuracy. They found robust correlations between centrilobular and panlobular emphysema types and various pulmonary metrics.

Moghadas-Dastjerdi *et al.* proposed an automated approach to evaluate COPD by scrutinizing different imaging modalities [23]. They extracted 23 features from CT scans of 69 COPD patients and investigated the correlation of their findings with the global initiative for chronic obstructive lung disease (GOLD) stage for diagnostic purposes. Polat *et al.* pioneered using a deep transfer learning network to assess COPD severity from chest CT images [24]. Their groundbreaking study achieved a high average accuracy of 96.79% in classifying moderate, and severe COPD, demonstrating the potential of transfer learning in expediting precise medical decisions.

Researchers have explored incorporating spirometry test data with supplementary datasets to evaluate the severity of COPD. The study conducted by Sun *et al.* utilized deep learning models and chest CT images from diverse datasets of Chinese hospitals [25]. Their attention-based multi-instance learning (MIL) models achieved an area under the curve (AUC) of 0.934 internally and 0.866 externally, enabling the automatic detection and staging of spirometry-defined COPD. According to the GOLD scale, a 3D residual network accurately graded 76.4% of patients. In another study, Haider *et al.* employed machine learning techniques to analyze respiratory sounds to differentiate between normal and COPD subjects [26]. Their classifiers, including support vector machine (SVM), k-nearest neighbors (KNN), logistic regression (LR), decision tree (DT), and discriminant analysis (DA), achieved up to 83.6% accuracy with SVM, relying on median frequency and linear predictive coeffcients. By combining lung sound and spirometry parameters, they achieved 100% accuracy, providing promising results for the future development of diagnostic systems and improvements in COPD diagnosis for clinicians' routine practices.

Several investigations have been conducted to explore the genetic factors associated with COPD, utilizing studies such as COPDGene. In this regard, Bhatt *et al.* have proposed a novel COPD severity classification known as STaging of Airflow Obstruction by Ratio (STAR), employing the FEV1/FVC ratio instead of ppFEV1 [27]. STAR has been validated across various datasets and has demonstrated robust agreement with GOLD staging, surpassing it in predicting critical outcomes such as mortality, quality of life, symptoms, and disease progression. It provides a more precise and less demographic-sensitive assessment of COPD severity compared to the ppFEV1-based GOLD system. Ying *et al.* have developed deep learning-based classifiers for exacerbation frequency and COPD severity, utilizing a three-layer deep belief network (DBN) on COPDGene data [28], [29]. They achieved 91.99% and 97.2% accuracy rates through ten-fold cross-validation, respectively. These studies have emphasized the significance of DBN in evaluating COPD severity and the risk of exacerbation.

Some other research in this field aimed to assess the severity of the disease using physiological data and laboratory findings. One such study by Swaminathan *et al.* proposed a machine learning approach that could detect early exacer bations of COPD, thereby mitigating patient risks [30]. The model, trained on physician opinions, demonstrated higher accuracy and safety in identifying and predicting exacerbations compared to individual pulmonologists. It exhibited potential as an effective decision-support tool for COPD management. Thomas Kronborg *et al.* compared one-layer and two-layer probabilistic models for predicting COPD exacerbations using continuous oxygen saturation, pulse rate, and blood pressure measurements [31]. Their study showed that the two-layer model significantly improved classification rates, with increased area under the ROC curve and higher sensitivity at a specificity of 0.95, demonstrating its clinical relevance for telehealth applications. Another study by Peng *et al.* addressed the lack of assessment methods for acute exacerbation in hospitalized COPD patients by developing a C5.0 decision tree classifier [32]. By analyzing data from 410 patients, including vital signs and inflammatory markers, the model achieved 80.3% accuracy in prognosis prediction, outperforming other models. This tool aids respiratory physicians in early severity assessment, guiding treatments, and enhancing patient outcomes during acute exacerbations of COPD.

Furthermore, Zheng *et al.* developed an assessment system for COPD severity to assist doctors in patient allocation across medical institutions [33]. Their HFL-COPRAS method incorporates clinical experience and system engineering, enabling decision-making in hesitant, fuzzy linguistic environments. The method was applied at the West China Hospital and assists in assessing COPD patient severity, thus facilitating appropriate hospital-level allocation based on severity assessment.

The studies reviewed in this section mainly employed methods that require many features to evaluate the severity of COPD. These methods typically involve analyzing CT scan images or spirometry results, which may not be feasible for ICU patients. However, the present study aims to take a different approach by utilizing the MIMIC-III dataset in a novel way. Specifically, this research uses machine learning algorithms to classify COPD severity by identifying the minimum number of features that can be easily measured in an ICU environment. This study seeks to develop a more practical and efficient method for assessing COPD severity in intensive care settings by concentrating on readily available and quickly obtainable data points.

## 3 Materials and Methods

### 3.1 Dataset

To conduct the research, we extracted the data from the MIMIC-III (Medical Information Mart for Intensive Care) dataset, accessible through the PhysioNet website [15]. MIMIC-III is a comprehensive database from a major tertiary care hospital encompassing a wide range of information about patients admitted to critical care units [16]. This information includes vital signs, medications, laboratory measurements, care providers' observations and notes, fluid balance, procedure codes, diagnostic codes, imaging reports, hospital length of stay, survival data, and other pertinent details.

### 3.2 Preprocessing

The acquisition of required data for this study necessitated intertable communication due to the expansive and interconnected nature of the MIMIC-III dataset. A procedural framework was developed to outline the setup and data retrieval process, as shown in Figure 1. During this process, we extracted crucial parameters. This task necessitated the establishment of connections between disparate tables to consolidate pertinent values into a single table. The resulting unified table was then subjected to additional processing steps, which involved linking data based on the Hospital admission ID (HADM-ID) and time in columnar format. The employed procedure included the following steps:

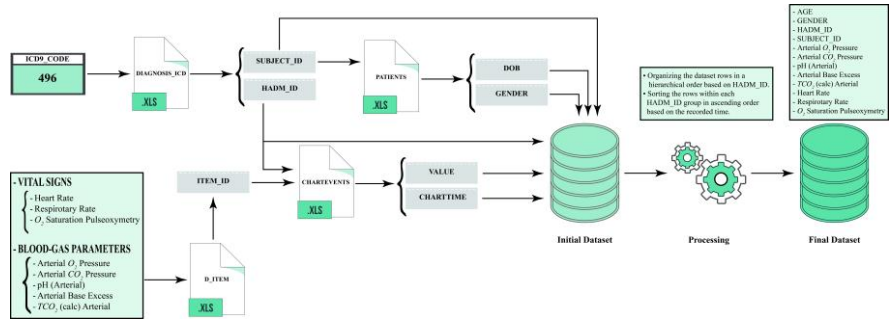

**Fig. 1** Data processing pipeline. The initial datasets, comprising diagnosis codes, vital signs, and blood-gas parameters, are extracted from various sources. These datasets are subsequently organized and processed to generate the final dataset, which includes the relevant features necessary for analysis.

1. Extracting ICUSTAY ID (Patient ICU stay ID) and HADM-ID from DIAGNOSES-ICD.csv (Diagnose International Classification of Diseases),
2. Identifying ITEMIDs for vital signs and blood gas parameters from D-ITEM.csv (Definition table for all items),
3. Extracting vital signs and blood gas parameters using HADM-ID in CHARTEVENTS.csv (All charted data for all patients),
4. Obtaining Age and Gender from demographics data,
5. Arranging the data into a table with 12,131 samples. The columns include two IDs (ICUSTAY ID and HADM ID), age, gender, the date and time of measurements, vital signs, and blood gas parameters.

This process involved the assembly of 12,131 samples. Each sample contained ten distinct attributes, as shown in Table 1. Measured parameters from the patients at different times were included in each sample. Additionally, each sample is characterized by two identifiers, ICUSTAY-ID and HADM-ID, along with a temporal marker corresponding to the parameter measurement instance. Notably, the algorithms exclusively consider the ten characteristics elucidated in Table 1.

### 3.3 Labeling

One limitation of the MIMIC-III dataset is the absence of pre-defined disease severity levels. This is why our work represents an innovative use of this dataset. The collected data lacked labels to determine the severity of COPD. Two pulmonologists were consulted to label some samples. Based on their assessment, an algorithm was created to satisfy both methodologies in categorizing the severity of the disease (Algorithm 1).

The normal range for the parameters is defined as follows:

- $7.35 \leq pH \leq 7.45$
- $54 \leq PO_2 \leq 67.6$
- $35 \leq PCO_2 \leq 45$
- $-3 \leq BE \leq 3$
- $23 \leq TCO_2 \leq 29$

**Table 1** VARIABLE UNITS AND TYPES

| Variable | Units |
|---|---|
| *Patient Profile* | |
| Age | year |
| Gender | male/female |
| *Blood Gas Parameters* | |
| Arterial $O_2$ pressure *(PO$_2$)* | mmHg |
| Arterial $CO_2$ Pressure *(PCO$_2$)* | mmHg |
| pH (Arterial) | None |
| Arterial Base Excess *(BE)* | None |
| Total Carbon Dioxide *(TCO$_2$)* | mmol/L |
| *Vital Signs* | |
| Heart Rate *(HR)* | bpm |
| Respiratory Rate *(RR)* | insp/min |
| $O_2$ saturation pulse oximetry *(SpO$_2$)* | % |

A sample is considered Mild to Moderate if:

- All parameters are within their normal range,
- *pH* is within the normal range, AND at least one of the following conditions is met:
  – PO$_2$ is within the normal range,
  – PCO$_2$ is within the normal range,
  – BE is within the normal range,
  –TCO$_2$ is within the normal range.

A sample is considered Severe if:

- All parameters are outside their normal range,
- *pH* is outside the normal range, AND at least one of the following conditions is met:
  – PO$_2$ is outside the normal range,
  – PCO$_2$ is outside the normal range,
  – BE is outside the normal range,
  – TCO$_2$ is outside the normal range.

## Labeling Guidelines

- If a sample meets the conditions for Severe, it should be labeled as 1.
- If a sample meets the conditions for Mild to Moderate, it should be labeled as 0.
- If a sample does not meet the Mild to Moderate or Severe conditions, it should not be labeled.

An algorithm was then implemented to classify the unlabeled samples in the subsequent stage following the medical professional's assessments into three groups: mild to moderate, severe, and unlabeled. Of all the samples available in the dataset, 3,282 were identified as mild to moderate, while 5,343 were classified as severe. However, 3,488 samples remained unlabeled and were not assigned to any group. We incorporated unlabeled data into our analysis to ensure that the algorithms do not simply replicate the patterns based on doctors' opinions and achieve 100% accuracy. Relying solely on labeled data could lead to overfitting, where the algorithms might identify patterns specific to the labeled set without true generalizability. We utilized unlabeled data to mitigate this and introduce uncertainty, initially labeled using semi-supervised algorithms. This approach enhances the robustness and generalizability of our findings. Semi-supervised learning algorithms were employed to assign labels to the unlabeled data to deal with this issue.

---

**Algorithm 1** Group Classification
---
1: $A \Leftarrow [7.35, 7.45]z$
2: $B \Leftarrow [54, 67.6]$
3: $C \Leftarrow [35, 45]$
4: $D \Leftarrow [-3, 3]$
5: $E \Leftarrow [23, 29]$
6: **for** each sample $s$ **do**
7:     **if** $s.pH \in A$ and $s.PO_2 \in B$ and $s.PCO_2 \in C$ and $s.BE \in D$ and $s.TCO_2 \in E$ **then**
8:         Display "$s$ is in the mild to moderate group"
9:     **else if** $s.pH \in A$ and ($s.PCO_2 \in C$ or $s.TCO_2 \in E$ or $s.BE \in D$) **then**
10:         Display "$s$ is in the mild to moderate group"
11:     **else if** $s.pH \notin A$ and $s.PO_2 \notin B$ and $s.PCO_2 \notin C$ and $s.BE \notin D$ and $s.TCO_2 \notin E$ **then**
12:         Display "$s$ is in the severe group"
13:     **else if** $s.pH \notin A$ and ($s.PCO_2 \notin C$ or $s.TCO_2 \notin E$ or $s.BE \notin D$) **then**
14:         Display "$s$ is in the severe group"
15:     **else**
16:         Display "$s$ is in the group without a label"
17:     **end if**
18: **end for**

---

### 3.4 Semi-Supervised Learning

After initial labeling, nearly one-third of the samples were left unlabeled. We employed semi-supervised algorithms, which are practical when dealing with datasets that have both labeled and unlabeled instances [34], [35]. We used two semi-supervised learning techniques, namely, the label spreading algorithm and the label propagation algorithm, to assign labels to the unlabeled samples. Label propagation computes a similarity matrix among samples using the k-nearest neighbors (KNN) methodology to propagate the labels [35]. On the other hand, label spreading follows the same process but includes a regularization mechanism to improve resistance against noise [36], [37]. The rationale for employing these two semi-supervised learning algorithms is as follows:

1. Semi-Supervised Learning Suitability: The dataset contained a mix of labeled and unlabeled samples, making it an ideal candidate for semi-supervised learning. These algorithms can effectively utilize the labeled data to predict the labels of the unlabeled data, thereby leveraging the information present in the entire dataset.
2. Medical Data Characteristics: The medical parameters in our dataset *(PO2, PCO2, pH, BE, HR, RR, SpO2, TCO2)* exhibit continuous values and are interrelated. Label propagation and label spreading algorithms are well-suited for such data as they assume that similar samples are likely to have similar labels, facilitating accurate label inference based on the proximity of data points in the feature space.
3. Robustness and Reliability: Applying both algorithms independently resulted in identical labeling outcomes, underscoring the robustness of our approach. This consistency across different algorithms enhances the reliability of the inferred labels and confirms these methods' appropriateness for our dataset.
4. Expert-Guided Labeling: The initial labeling process was guided by a flowchart developed in consultation with lung doctors, ensuring that the criteria for severity classification were medically sound. The semi-supervised algorithms extended this expert-guided labeling, maintaining the integrity and validity of the classification process.
5. Data Structure Preservation: These algorithms respect the inherent structure of the data, preserving local and global relationships between samples. This is particularly important in medical datasets, where slight parameter variations can significantly impact disease severity; thus, accurate label propagation is crucial.

After this process, out of the initial 3,488 unlabeled samples, 2,099 were categorized as mild to moderate, while 1,389 samples fell under the severe category. This new categorization resulted in a dataset comprising 5,381 samples under the mild to moderate class and 6,729 samples under the severe class.

### 3.5 Classification Algorithms

We utilized three distinct machine learning models, random forest, support vector machine, and k-nearest neighbors, to classify the severity of the disease into two categories: mild to moderate and severe. All models followed a consistent methodological structure. We conducted data preprocessing, which involved the imputation of missing values and feature scaling to ensure uniformity across variables. For model evaluation, we employed a cross-validation strategy to assess the generalizability of our models. Through this systematic approach, we aimed to conduct a thorough comparison of the three machine learning models, providing insights into their relative strengths and weaknesses for our specific classification task. We use SimpleImputer to fill missing values with column means and apply StandardScaler to normalize the data for each three algorithms.

1. **Random Forest**: Our study employed the random forest algorithm, an ensemble learning method that leverages multiple decision trees to enhance accuracy and mitigate overfitting [38]. Its capability to handle complex interactions and provide feature-importance insights makes it ideal for medical data, where interpretability and reliability are crucial. Recognizing that using the random forest algorithm without proper hyperparameter adjustment can lead to suboptimal performance, we conducted hyperparameter tuning to optimize its performance as our foundational classifier. As a starting point, we initialized the random forest Classifier with 100 trees (n_estimators=100) and a maximum depth of 10 (max_depth=10). These initial values were selected to balance model complexity and computational efficiency. To ensure the reproducibility of our results, we set the random_state parameter to 42. This configuration served as our baseline, from which we further refined the model through systematic hyperparameter optimization to achieve the best possible performance for our specific classification task.

2. **KNN**: The k-nearest neighbors (KNN) algorithm is a non-parametric method for classification and regression tasks. It operates on the principle of similarity, classifying new data points based on the majority class of their k-nearest neighbors in the feature space. While KNN is intuitive and easy to implement, its performance can be susceptible to the choice of k and the distance metric used [39]. In our implementation, we employed a systematic approach to determine the optimal number of neighbors (k) for our KNN model. We evaluated odd values of k ranging from 1 to 29, using cross-validated accuracy as our performance metric. The KNN classifier was initialized with uniform weights, meaning all neighbors contribute equally to the classification decision. We set the randomstate to 42 for our cross-validation splits to ensure reproducibility. This methodical search for the optimal k value allowed us to find the best trade-off between underfitting and overfitting, tailoring the model to the specific characteristics of our dataset. Once the optimal k was determined, we used this value to train our final KNN model and evaluate its performance across various metrics.

3. **SVM**: Support vector machine (SVM) is a robust supervised learning algorithm for classification and regression tasks. It works by finding the hyperplane that best separates different classes in a high-dimensional space, maximizing the margin between classes. SVMs are particularly effective in high-dimensional spaces and are versatile due to their ability to use different kernel functions [40]. In our implementation, we utilized the SVM classifier with its default parameters, except for enabling probability estimates by setting probability=True. This allows the model to output probability scores for each class, which can be helpful in more nuanced decision-making and model evaluation. We did not perform explicit hyperparameter tuning in this initial setup; instead, we used the default configuration as a baseline. The random state was set to 42 to ensure the reproducibility of our results. We employed a 5-fold stratified cross-validation strategy to robustly evaluate the model's performance across various metrics. This approach provided a comprehensive assessment of the SVM's performance on our specific dataset, serving as a foundation for potential future optimizations.

## 4 Results

The dataset comprises ten feature columns utilized as inputs for the algorithms shown in Table 1, along with a label column indicating disease severity. Initially, missing values in the dataset were filled with column means using the SimpleImputer strategy.

Subsequently, the data underwent normalization using StandardScaler before being fed into the algorithms. After these preprocessing steps, we ran three separate random forest algorithms, k-nearest neighbors (KNN), and support vector machine (SVM) on the data to evaluate the severity of the disease. To assess the performance of each model, a Stratified k-fold cross-validation approach with five folds was employed. We utilized a range of performance metrics, including accuracy, precision, recall, F1 score, and area under the receiver operating characteristic curve (ROC AUC). These metrics were calculated for both training and test sets to provide a comprehensive view of model performance and to detect any potential overfitting. To enhance our understanding of model behavior and facilitate parameter selection, we incorporated various visualization techniques. These visual aids served to illustrate model-specific characteristics, such as feature importance for random forest, or to guide the selection of optimal hyperparameters, as in the case of k-nearest neighbors. The random forest classifier exhibited excellent performance, achieving a test accuracy of 0.9251 and a ROC AUC of 0.9841. The k-nearest neighbors (KNN) classifier, with an optimal number of neighbors set to 17, demonstrated solid but comparatively lower performance with a test accu- racy of 0.8816 and a ROC AUC of 0.9583. The support vector machine (SVM) classifier provided robust metrics, achieving a test accuracy of 0.9039 and a ROC AUC of 0.9721. Complete results are detailed in Table 2. Additionally, the confusion matrix and the AUC curve are illustrated in Figure 2 and Figure 3, respectively. According to the above results, the random forest classification delivers the best performance among the evaluated models. With the highest test accuracy and ROC AUC, the random forest classifier proves to be the most effective in classifying the severity of COPD in this dataset. The SVM and KNN classifiers, while demonstrating strong results, do not surpass the accuracy and reliability of the random forest model, making it the most suitable algorithm for this task.

**Table 2** Classifier Performance Metrics

| Classifier | Accuracy | Precision | Recall | F1 | ROC AUC |
|---|---|---|---|---|---|
| Random Forest | 0.9251 (± 0.0105) | 0.9569 (± 0.0114) | 0.9061 (± 0.0136) | 0.9308 (± 0.0099) | 0.9841 (± 0.0030) |
| KNN (k=17) | 0.8816 (± 0.0117) | 0.9151 (± 0.0160) | 0.8675 (± 0.0127) | 0.8907 (± 0.0105) | 0.9583 (± 0.0070) |
| SVM | 0.9039 (± 0.0117) | 0.9362 (± 0.0127) | 0.8876 (± 0.0090) | 0.9112 (± 0.0107) | 0.9721 (± 0.0041) |

# 5 Discussion

This study demonstrates the effectiveness of machine learning algorithms, particularly random forest, in classifying COPD severity using readily available ICU data from the MIMIC-III dataset. Our approach achieves high accuracy and ROC AUC using only ten features derived from blood gas parameters and vital signs without relying on more complex data like CT scans or spirometry results. This represents a significant step forward in applying machine learning to COPD management in intensive care settings. Our work stands out in its novel use of the MIMIC-III dataset for COPD severity classification. While many previous studies have utilized this dataset for mortality prediction or other diagnostic purposes, our application extends its utility into disease severity assessment. This innovative approach allows us to leverage the rich, real-world data in MIMIC-III to address a critical clinical need in COPD management. A key strength of our methodology lies in its simplicity and accessibility. Unlike studies that rely on CT imaging or spirometry data, our approach uses only parameters routinely collected in ICU settings. This makes our model particularly suited for rapid assessment of critically ill patients, where more complex diagnostic procedures may need to be more practical and possible. For instance, Sun *et al.* [25] achieved impressive results using deep learning models on chest CT images. However, our method offers a more immediately applicable solution in ICU environments where CT scans may not be readily available or advisable for all patients.

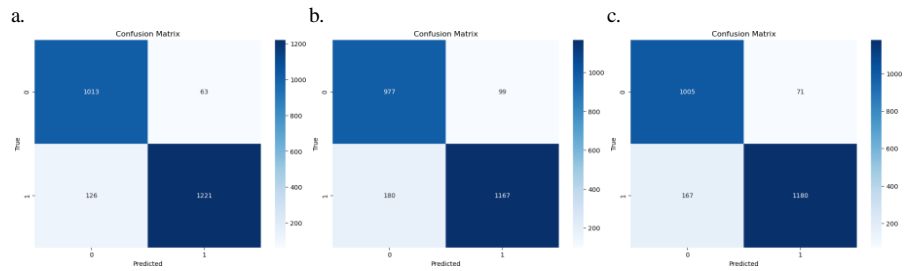

**Fig. 2** Confusion matrices representing the performance of three different models: (a) random forest classifier, (b) KNN classifier, and (c) SVM classifier.

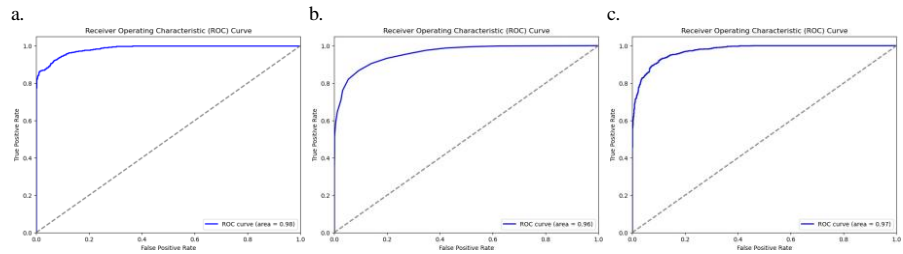

**Fig. 3** ROC curve representing the performance of three different models: (a) random forest classifier, (b) KNN classifier, and (c) SVM classifier.

The minimal feature set required by our model is another significant advantage. We achieved high accuracy with only ten features, unlike studies like that of Peng *et al.* [32], which incorporated a more extensive set of variables, including inflammatory markers. Our streamlined approach simplifies data collection and enhances the model's potential for real-time application in clinical settings. Our use of semi-supervised learning techniques to label additional data represents a novel approach in this field. By effectively leveraging unlabeled data, we have enhanced the robustness of our model and made more efficient use of the available information in the dataset. This method could prove valuable in other medical machine learning applications where labeled data is scarce, but unlabeled data is plentiful. In terms of performance, our random forest model demonstrates excellent accuracy, outperforming some previous COPD classification efforts. For example, the decision tree classifier developed by Peng *et al.* [32] for assessing acute exacerbation in hospitalized COPD patients achieved 80.3% accuracy, while our model reached higher levels of accuracy. However, it is essential to note that direct comparisons are challenging due to differences in datasets and specific classification tasks. Our approach offers more immediate clinical applicability when compared to genetic studies like those of Bhatt *et al.* [27] and Ying *et al.* [28], [29]. While genetic factors undoubtedly play a crucial role in COPD, our model's reliance on readily available clinical parameters makes it more suitable for real-time decision-making in ICU settings. That said, future integration of genetic data could enhance our model's predictive power.

Our results align with and build upon other studies that have shown the value of machine learning in COPD assessment. For instance, the work of Swaminathan *et al.* [30] in detecting early exacerbations and Thomas Kronborg *et al.* [31] in using probabilistic models for predicting exacerbations both highlight the potential of machine learning in COPD management. Our study extends this potential into severity classification, offering a complementary tool for comprehensive COPD care. The approach developed by Zheng *et al.* [33] for COPD severity assessment to assist in patient allocation across medical institutions shares some similarities with our work in its goal of aiding clinical decision-making. However, our focus on ICU-specific parameters and the use of the MIMIC-III dataset distinguishes our approach and makes it particularly suited for critical care settings. It is important to acknowledge our study's limitations. The labeling process, while guided by expert opinion, introduces some subjectivity. Additionally, our model's performance on external datasets remains to be validated. Future work should focus on external validation and explore how this approach could be implemented in clinical decision-support systems. In conclusion, our study presents a novel, efficient, and highly accurate method for COPD severity classification in ICU settings. By leveraging machine learning techniques and readily available clinical data, we have developed a tool that has the potential to enhance COPD management in critical care environments significantly. The simplicity and accessibility of our approach, combined with its high performance, make it a promising candidate for integration into clinical practice, potentially leading to more timely and appropriate interventions for COPD patients in intensive care.

## 6 Conclusion

We proposed a practical approach to COPD severity classification using machine learning algorithms applied to the MIMIC-III dataset. By focusing on a minimal set of readily available ICU parameters, we developed a model that achieves high accuracy and ROC AUC, distinguishing between mild-to-moderate and severe COPD cases. This work represents an advancement in the application of machine learning to critical care management of COPD patients. We efficiently used the ICU data for COPD severity classification, achieving high accuracy using ten features derived from blood gas parameters and vital signs. This approach is valuable in ICU settings where rapid, non-invasive assessment is crucial. We also proposed comprehensive pre-processing and a novel application of the MIMIC-III dataset, extending its utility beyond traditional applications such as mortality prediction and showcasing its potential for disease severity classification. Implementing semi-supervised learning techniques to utilize unlabeled data effectively is a significant methodological contribution, enhancing our model's robustness and demonstrating a way to maximize the value of clinical datasets where labelled data may be limited.

The high performance of our random forest model, combined with its reliance on easily accessible parameters, makes it highly practical for clinical use. This balance of accuracy and accessibility could facilitate more widespread adoption of machine learning tools in critical care settings. The ability to quickly and accurately assess COPD severity using readily available parameters could lead to more timely and appropriate interventions, potentially improving patient outcomes. Furthermore, this approach could help standardize COPD severity assessment across different ICU settings, leading to more consistent care. However, this study also points to several areas for future research, including external validation of the model on diverse datasets, integration of additional data types, and exploring the model's capability to predict COPD exacerbations or other clinically relevant outcomes. Investigating how this approach could be implemented in clinical decision-support systems is also crucial. As we continue to refine and validate this approach, it has the potential to become a valuable asset in the ongoing effort to improve outcomes for patients with severe COPD.

supervised deep learning approach. European Radiology, 32(8), 5319–5329. https://doi.org/10.1007/s00330-022-08632-7

[26] Haider, N. S., Singh, B. K., Periyasamy, R., & Behera, A. K. (2019). Respiratory Sound Based Classification of Chronic Obstructive Pulmonary Disease: a Risk Stratification Approach in Machine Learning Paradigm. Journal of Medical Systems, 43(8). https://doi.org/10.1007/s10916-019-1388-0

[27] Bhatt, S. P., Nakhmani, A., Fortis, S., Strand, M. J., Silverman, E. K., Sciurba, F. C., & Bodduluri, S. (2023). FEV1/FVC Severity Stages for Chronic Obstructive Pulmonary Disease. American Journal of Respiratory and Critical Care Medicine, 208(6), 676–684. https://doi.org/10.1164/rccm.202303-0450OC

[28] Ying, J., Dutta, J., Guo, N., Hu, C., Zhou, D., Sitek, A., & Li, Q. (2020). Classification of Exacerbation Frequency in the COPDGene Cohort Using Deep Learning With Deep Belief Networks. IEEE Journal of Biomedical and Health Informatics, 24(6), 1805–1813. https://doi.org/10.1109/jbhi.2016.2642944

[29] Ying, J., Dutta, J., Guo, N., Xia, L., Sitek, A., & Li, Q. (2016). Gold classification of COPDGene cohort based on deep learning. https://doi.org/10.1109/icassp.2016.7472122

[30] Swaminathan, S., Qirko, K., Smith, T., Corcoran, E., Wysham, N. G., Bazaz, G., Kappel, G., & Gerber, A. N. (2017). A machine learning approach to triaging patients with chronic obstructive pulmonary disease. PloS One, 12(11), e0188532. https://doi.org/10.1371/journal.pone.0188532

[31] Kronborg, T., Hangaard, S., Cichosz, S. L., & Hejlesen, O. (2021). A two-layer probabilistic model to predict COPD exacerbations for patients in telehealth. Computers in Biology and Medicine, 128, 104108. https://doi.org/10.1016/j.compbiomed.2020.104108

[32] Peng, J., Chen, C., Zhou, M., Xie, X., Zhou, Y., & Luo, C.-H. (2020). A Machine-learning Approach to Forecast Aggravation Risk in Patients with Acute Exacerbation of Chronic Obstructive Pulmonary Disease with Clinical Indicators. Scientific Reports, 10(1), 3118. https://doi.org/10.1038/s41598-020-60042-1

[33] Zheng, Y., Xu, Z., He, Y., & Liao, H. (2018). Severity assessment of chronic obstructive pulmonary disease based on hesitant fuzzy linguistic COPRAS method. Applied Soft Computing, 69, 60–71. https://doi.org/10.1016/j.asoc.2018.04.035

[34] Hands-on Machine Learning with Scikit-Learn, Keras, and TensorFlow, 2nd Edition. (2019). Hands-on Machine Learning with Scikit-Learn, Keras, and TensorFlow, 2nd Edition. O'Reilly — Safari. https://www.oreilly.com/library/view/hands-on-machine-learning/9781492032632/